\documentclass[conference]{IEEEtran}
\IEEEoverridecommandlockouts
\usepackage{cite}
\usepackage{amsmath,amssymb,amsfonts}
\usepackage{algorithmic}
\usepackage{graphicx}
\usepackage{textcomp}
\usepackage{xcolor}

\usepackage{url}
\usepackage{booktabs}
\usepackage{colortbl}
\definecolor{mygray}{gray}{0.9}
\definecolor{tablegray}{gray}{.9}
\definecolor{tableblue}{RGB}{221,238,251}
\definecolor{tablegreen}{RGB}{225,245,245}
\definecolor{baselinecolor}{gray}{.9}
\newcommand{\baseline}[1]{\cellcolor{baselinecolor}{#1}}
\usepackage[colorlinks = true,
    linkcolor = black,
    urlcolor  = blue,
    citecolor = magenta,
    bookmarks = true, pagebackref]{hyperref}

\newcolumntype{x}[1]{>{\centering\arraybackslash}p{#1pt}}
\newcolumntype{y}[1]{>{\raggedright\arraybackslash}p{#1pt}}
\newcolumntype{z}[1]{>{\raggedleft\arraybackslash}p{#1pt}}

\usepackage{array}
\usepackage{stfloats}
\usepackage{verbatim}
\usepackage{color}
\usepackage{multirow}
\usepackage{bbm}
\usepackage{bm}
\usepackage{makecell}
\usepackage{gensymb}
\def\BibTeX{{\rm B\kern-.05em{\sc i\kern-.025em b}\kern-.08em
    T\kern-.1667em\lower.7ex\hbox{E}\kern-.125emX}}
\begin{document}

\title{ESVQA: Perceptual Quality Assessment of Egocentric Spatial Videos}

\author{\textit{Xilei Zhu}$^1$, \textit{Huiyu Duan}$^{1 \dagger}$, \textit{Liu Yang}$^1$, \textit{Yucheng Zhu}$^1$, \textit{Xiongkuo Min}$^{1 \dagger}$, \textit{Guangtao Zhai}$^{1 \dagger}$, \textit{Patrick Le Callet}$^2$\\
$^1$Institute of Image Communication and Network Engineering, Shanghai Jiao Tong University\\
$^2$Institut Universitaire de France (IUF), University of Nantes\\
\tt\small \{xilei\_zhu, huiyuduan, ylyl.yl, zyc420, minxiongkuo, zhaiguangtao\}@sjtu.edu.cn, \\ 
\tt\small patrick.lecallet@univ-nantes.fr \\
\thanks{$^{\dagger}$Corresponding authors: Huiyu Duan, Xiongkuo Min, and Guangtao Zhai.}
\thanks{This work was supported in part by the National Natural Science Foundation of China under Grants 62401365, 62225112, 62271312, 62132006.}
}

\vspace{-15pt}

\maketitle

\begin{abstract}
With the rapid development of eXtended Reality (XR), egocentric spatial shooting and display technologies have further enhanced immersion and engagement for users, delivering more captivating and interactive experiences.
Assessing the quality of experience (QoE) of egocentric spatial videos is crucial to ensure a high-quality viewing experience. 
However, the corresponding research is still lacking.
In this paper, we use the concept of \textit{embodied experience} to highlight this more immersive experience and study the new problem, \textit{i.e.,} embodied perceptual quality assessment for egocentric spatial videos.
Specifically, we introduce the first \underline{\textbf{E}}gocentric \underline{\textbf{S}}patial \underline{\textbf{V}}ideo \underline{\textbf{Q}}uality \underline{\textbf{A}}ssessment \underline{\textbf{D}}atabase (ESVQAD), which comprises 600 egocentric spatial videos captured using the Apple Vision Pro and their corresponding mean opinion scores (MOSs). Furthermore, we propose a novel multi-dimensional binocular feature fusion model, termed ESVQAnet, which integrates binocular spatial, motion, and semantic features to predict the overall perceptual quality. Experimental results demonstrate the ESVQAnet significantly outperforms 16 state-of-the-art VQA models on the embodied perceptual quality assessment task, and exhibits strong generalization capability on traditional VQA tasks. The database and code are available at \textcolor{magenta}{https://github.com/iamazxl/ESVQA}.
\end{abstract}


\section{Introduction}
Egocentric shooting and display technologies have gained widespread attention recently, providing more convenient capturing methods and immersive viewing experiences through head-mounted displays (HMDs) such as Apple Vision Pro, Meta Quest 3, and Google Glass \cite{min2024perceptual,zhu2024esiqa}. 
The captured first-person binocular perspective videos, \textit{i.e.,} egocentric spatial videos, can reproduce the physical and sensory experience of the shooters, which significantly improves immersion and engagement for viewers.
However, the egocentric spatial videos taken by a wide variety of individuals from professional photographers to amateurs, exhibit significant differences in quality of experience (QoE) \cite{min2024perceptual, zhu2024esiqa, sun2022deep}, which is further influenced by the dynamic and uncontrolled capture environments  \cite{li2022blindly}, leading to common distortions such as under/overexposure, low visibility, noise, and color shifts, \textit{etc}.
Moreover, the QoE of egocentric spatial videos viewing in an immersive manner on HMDs is a more challenging problem, and the corresponding research is still lacking.

Over the past few decades, many studies have focused on the traditional 2D video quality assessment research, with establishing several VQA databases \cite{duan2024finevq, ying2021patch, hosu2017konstanz, wang2019youtube}, and developing advanced VQA models \cite{duan2024finevq, sun2022deep, wu2022fast, wu2022fasterquality}. 
However, these UGC VQA models are inadequate for assessing the quality of egocentric spatial videos due to their unique immersive and stereoscopic experience.
With the advancement of video media, some studies have explored the VR VQA problem \cite{zhu2023perceptual,duan2022confusing,duan2023attentive}. However, these studies only consider the immersive experience while ignore the stereoscopic effects. Existing 3D VQA studies \cite{8085198,8937476} are designed for traditional stereoscopic videos, which rely on dual-view to create depth perception. However, these metrics overlook the immersive experience in egocentric spatial videos, which feature a first-person perspective in dynamic environments. 
To the best of our knowledge, there is still no dedicated research on the quality assessment of egocentric spatial videos, highlighting the necessity for studies in this area.

In this paper, to further highlight the protagonist experience presented by viewing egocentric spatial videos in an immersive manner, we introduce the concept of \textit{embodied experience} from embodied artificial intelligence (AI) and study the embodied perceptual quality assessment problem of egocentric spatial videos.
To better understand human visual preferences for egocentric spatial videos and support the development of tailored VQA models, we establish the first \underline{\textbf{E}}gocentric \underline{\textbf{S}}patial \underline{\textbf{V}}ideo \underline{\textbf{Q}}uality \underline{\textbf{A}}ssessment \underline{\textbf{D}}atabase (ESVQAD), which consists of 600 egocentric spatial videos with diverse scenes, accompanied by the corresponding human perceptual quality ratings. We use Apple Vision Pro for both the egocentric video capturing process and the subjective evaluation procedure.
Based on the ESVQAD, we propose a novel multi-dimensional binocular feature fusion model, named ESVQAnet, which extracts spatial, motion, and semantic features from binocular views and then integrates to predict the overall quality score. 
The spatial feature extractor utilizes visual state space duality (VSSD) blocks \cite{shi2024vssd} and multi-head self-attention (MSA) blocks to hierarchically extract spatial features from binocular views.
The motion feature extractor leverages 3D-CNN to capture motion distortions and dynamic changes. The semantic feature extractor utilizes cascaded vision transformer layers of the CLIP visual encoder to extract high-level semantic information from the videos. 
Finally, the feature fusion module combines visual, semantic, and motion features to generate the quality score by a multi-layer perceptron (MLP). Extensive experimental results demonstrate that ESVQAnet significantly outperforms the benchmark models on ESVQAD and exhibits strong generalization performance on traditional VQA tasks.
Our contributions are summarized as follows:
\begin{itemize}
\vspace{-2pt}
\item  We introduce ESVQAD, the first perceptual quality assessment database dedicated to egocentric spatial videos to the best of our knowledge.  
\item We conduct a comprehensive benchmark experiment by evaluating the performance of numerous state-of-the-art VQA models based on ESVQAD.  
\item We propose a novel egocentric spatial video quality assessment model based on the multi-dimensional binocular feature fusion method, which significantly
 outperforms benchmark models in predicting perceptual quality.  \vspace{-2pt}
\end{itemize}

\section{Database Construction}\label{database}
\subsection{Content Collection}
\vspace{-2pt}


We use the Apple Vision Pro to capture egocentric spatial videos across various activities and scenes, highlighting diverse characteristics such as brightness, contrast, colorfulness, and spatial details. The raw videos are captured by 8 photographers, reflecting diverse user perspectives and preferences.
From the captured egocentric spatial videos, we select 200 high-quality samples as source videos for our ESVQAD database. Each selected video is standardized to a duration of 8 seconds, presented at a resolution of 2200$\times$2200 with a frame rate of 30 fps. The accompanying stereo audio offers high fidelity with a sampling rate of 48,000 Hz.
To simulate typical video quality degradations in streaming applications, we generate distorted versions of each source video. Using multiview high-efficiency video coding (MV-HEVC) via the Spatial Video Tool, each source video is compressed at three bitrates, \textit{i.e.,} 30 Mbps, 5 Mbps, and 1 Mbps, resulting in 600 egocentric spatial videos.



\begin{figure}
\vspace{-15pt}
    \centering
    \includegraphics[width=0.5\textwidth]{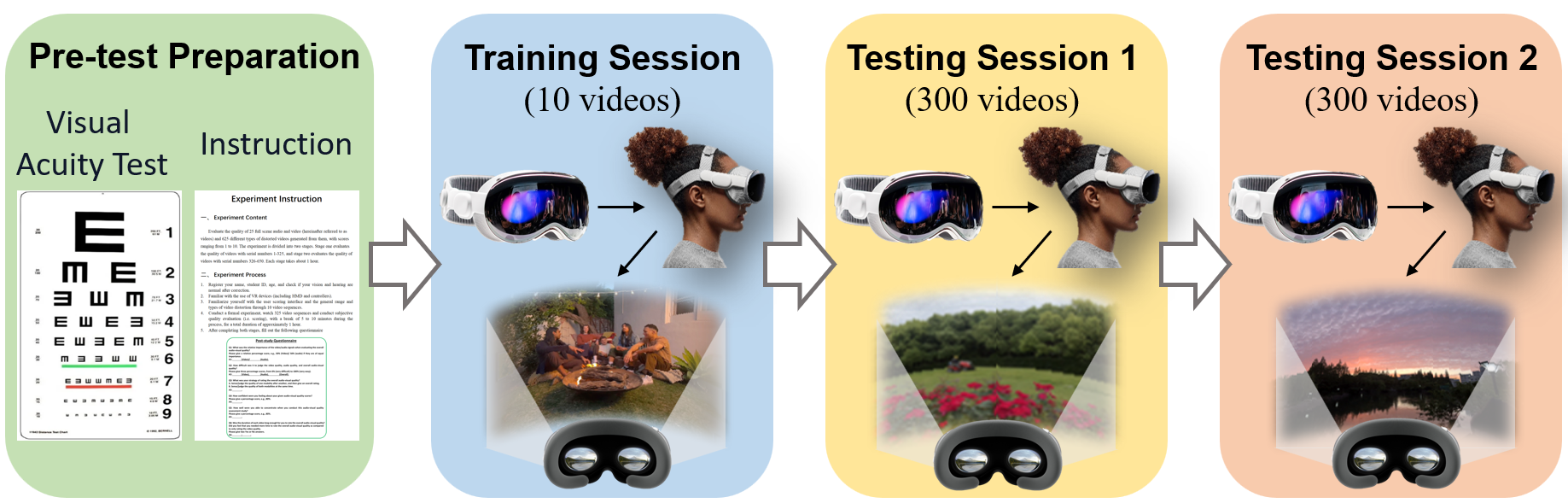}
    \vspace{-20pt}
    \caption{The workflow of our human subjective study for the ESVQAD database, consisting of 4 stages, \textit{i.e.,} a preparation session, a training session, and 2 testing sessions.}
    \label{workflow}
      \vspace{-16pt}
\end{figure}

\vspace{-4pt}
\subsection{Subjective Experiment}
\vspace{-2pt}
\subsubsection{Experiment setup}

We conduct a subjective experiment to collect human perceptual quality scores for the 600 egocentric spatial videos in a 3D immersive environment. We use the Apple Vision Pro as the HMD, which features a resolution of 3660$\times$3200 per eye, a 120$\degree$ horizontal field of view (FoV), and a 100Hz refresh rate. The videos are presented in a wide immersive FoV to ensure an authentic egocentric experience. All videos are displayed through the built-in ``Photos" app at their native resolutions to avoid scaling distortion.

\subsubsection{Experiment methodology}

We adopt the single stimulus absolute category rating (SSACR) method to collect subjective quality ratings for each video in ESVQAD, with the scale from 1 to 10. In the subjective experiments, participants are instructed to wear the Apple Vision Pro and watch videos through the ``Photos" app. A keyboard is placed in front of them to enter the scores. After viewing each video, participants manually enter their scores and proceed to the next video using gestures. To eliminate bias, the video order is randomized for each participant.
A total of 20 graduate students (10 males and 10 females) participate in the subjective experiment, which follows a four-stage workflow, as illustrated in Fig. \ref{workflow}. Before the experiment, all participants complete a standard visual assessment procedure using the Snellen visual acuity test and are confirmed to have normal or corrected-to-normal vision. 
They are instructed to evaluate egocentric spatial videos based on distortion and aesthetic quality, providing an overall quality score for each video. Each participant completes a training session and two testing sessions, producing a total of 12,000 quality ratings (20 participants $\times$ 600 ratings).

\vspace{-4pt}
\subsection{Subjective Data Processing and Analysis}
\vspace{-2pt}
We follow the subjective data processing guidelines recommended by ITU \cite{ITU-R_BT_500-13} for outlier detection and subject rejection. 
None of the 20 participants is considered an outlier and excluded.
The raw scores provided by the participants are normalized into Z-scores ranging from 0 to 100. Then we calculate the average of these Z-scores to derive the mean opinion scores (MOSs), which are formulated as follows: 
\begin{equation}
\setlength{\abovedisplayskip}{2pt}
\setlength{\belowdisplayskip}{0pt}
    z_{i j}  =\frac{r_{i j}-\mu_i}{\sigma_i}, \quad z_{i j}^{\prime}=\frac{100\left(z_{i j}+3\right)}{6},
    \vspace{-2pt}
\end{equation}
\begin{equation}
\setlength{\abovedisplayskip}{-3pt}
\setlength{\belowdisplayskip}{0pt}
    \text{MOS}_j  =\frac{1}{N} \sum_{i=1}^N z_{i j}^{\prime},
    \vspace{-4pt}
\end{equation}
where $r_{ij}$ is the original score of the $i$-th subject on the $j$-th video, $\mu_i$ and $\sigma_i$ represent the mean rating and the standard deviation given by subject $i$, and $N$ is the number of subjects.

\begin{figure}
\vspace{-20pt}
    \centering
    \includegraphics[width=0.43\textwidth]{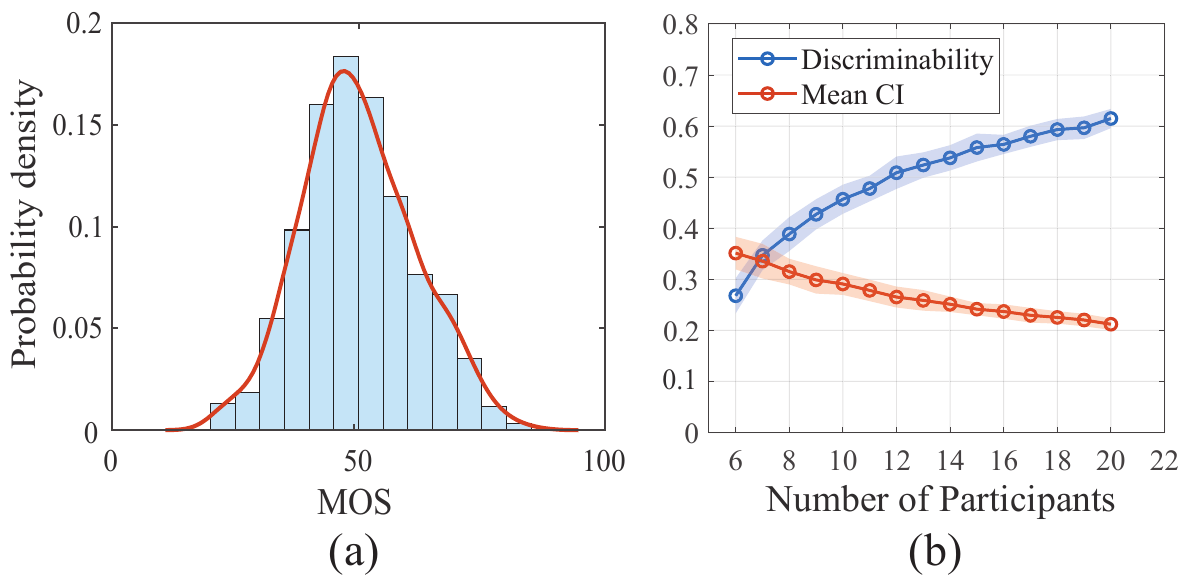} \vspace{-10pt}
    \caption{MOS analysis for the ESVQAD database. (a) Histogram of MOSs. (b) Evolution of MOS discriminability and mean confidence interval (CI) with increasing participant numbers. }
    \label{MOS}
      \vspace{-15pt}
\end{figure}

Fig. \ref{MOS}(a) shows the histogram of the MOS distribution, demonstrating that the perceptual quality scores are well-distributed across the [0, 100] range. To evaluate the reliability of MOSs, we calculate the \textit{discriminability} and \textit{mean confidence interval (CI)} metrics for the ESVQAD as the number of participants increased \cite{10448123}.
For the discriminability metric, we apply the two-sample Wilcoxon test to all possible MOS pairs in the ESVQAD, determining the proportion of significantly different pairs \cite{10448123}. The mean CI is calculated by averaging the standard deviations of the scores for each sample, scaled using a Z-score for a 95\% confidence level, to quantify the average uncertainty around the MOSs.
Fig. \ref{MOS}(b) illustrates the trends of both metrics as the number of participants increases, indicating that 20 participants provide relatively reliable MOSs.

\begin{figure}
\vspace{-10pt}
    \centering
\includegraphics[width=0.475\textwidth]{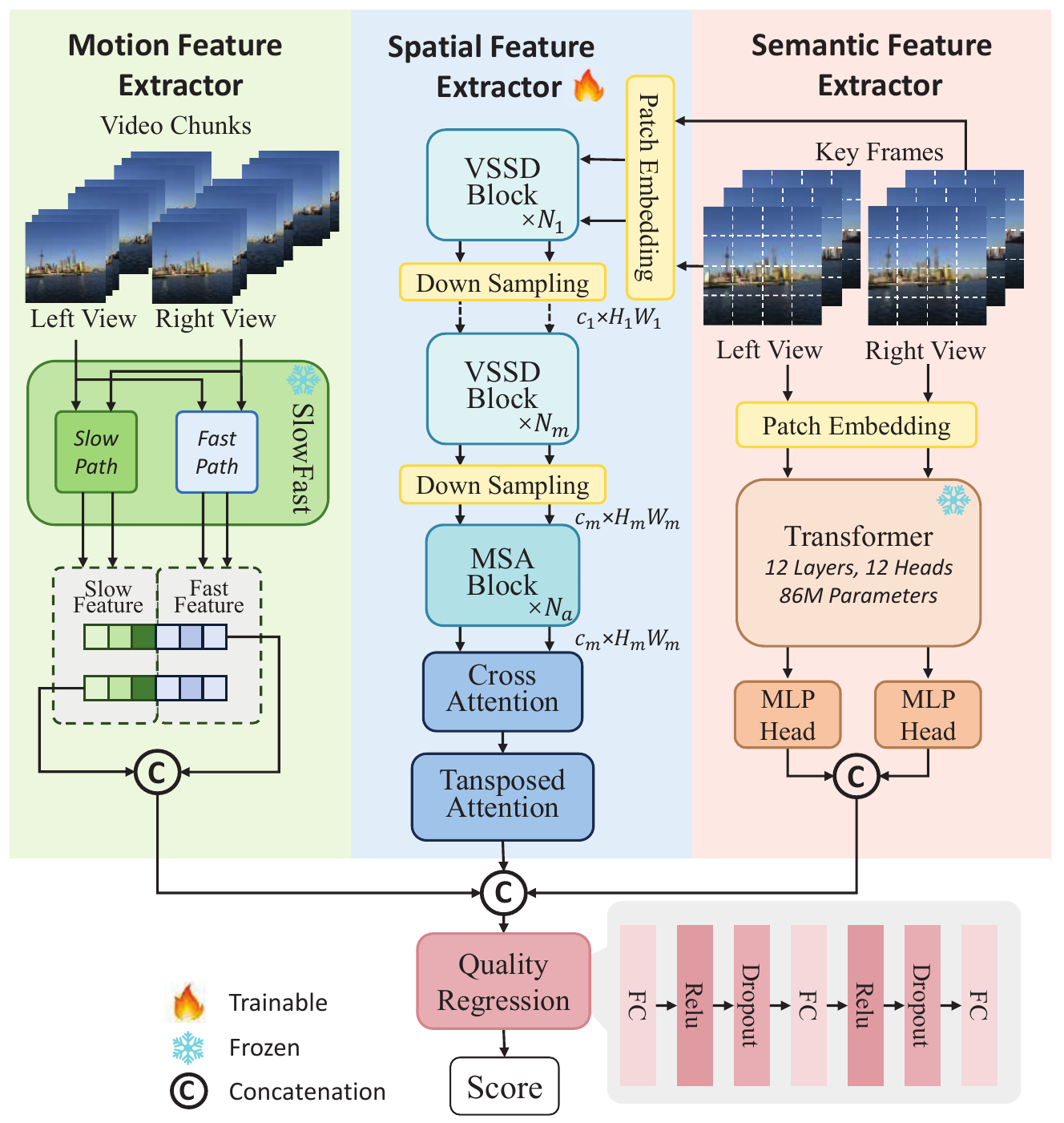} \vspace{-12pt}
    \caption{Illustration of the overall architecture of the proposed multi-dimensional binocular feature fusion model ESVQAnet, which consists of spatial, motion, and semantic feature extractors and one feature fusion module. }
    \label{framework}
    \vspace{-18pt}
\end{figure}

 \section{Proposed Method}
 \vspace{-3pt}
This section introduces the proposed \textbf{E}gocentric \textbf{S}patial \textbf{V}ideo \textbf{Q}uality \textbf{A}ssessment \textbf{Net}work, termed ESVQAnet, which consists of three feature extractors, \textit{i.e.,} spatial, motion, and semantic feature extractors, and one feature fusion module, as illustrated in Fig. \ref{framework}.
The spatial feature extractor employs VSSD blocks in the early stages and MSA blocks in the final stage, followed by cross-attention to fuse binocular view information. The motion feature extractor utilizes 3D-CNNs to model motion distortions and dynamic changes. The semantic feature extractor uses cascaded vision transformer layers to capture high-level semantic features. Finally, the feature fusion module combines spatial, motion and semantic features, generating the final quality score through an MLP.

\vspace{-6pt}
\subsection{Spatial Feature Extractor}
\vspace{-4pt}
 
In our model, we employ multi-stage of VSSD blocks \cite{shi2024vssd} and a final stage of MSA blocks to hierarchically extract spatial features from the binocular views of egocentric spatial video frames. 
The VSSD modules efficiently capture global visual features of video frames, progressively refining visual information through multiple stages from shallow to deep representations.
We decompose the video into two sequences representing the left and right views, denoted as $\boldsymbol{x}^l=\left\{\boldsymbol{x}^l_i\right\}_{i=0}^{N-1}$ and $\boldsymbol{x}^r=\left\{\boldsymbol{x}^r_i\right\}_{i=0}^{N-1}$, where $\boldsymbol{x}^l_i$ and $\boldsymbol{x}^r_i$ $\in$ $\mathbb{R}^{H \times W \times 3}$ corresponds to the $i$-th frame of the left and right views, respectively, $H$ and $W$ represent the spatial resolution, and $N$ is the total number of frames. 
To focus on key temporal features, we sample $N_t$ key frames from the video, which are denoted as $\boldsymbol{y}^l=\left\{\boldsymbol{y}^l_i\right\}_{i=0}^{N_t-1}$ and $\boldsymbol{y}^r=\left\{\boldsymbol{y}^r_i\right\}_{i=0}^{N_t-1}$. 
The $N_t$ key frames are processed through a patch embedding layer to generate feature maps $\textbf{\textit{F}}^{\textit{l}}_0$ and $\textbf{\textit{F}}^{\textit{r}}_0 \in \mathbb{R}^{N_t \times c_0 \times H_0 W_0}$. The feature maps are then fed into $m$ stages of VSSD blocks, where the \(i\)-th stage consists of \(N_i\) VSSD blocks, where \(i \in \{1, 2, \dots, m\}\).
In the \(i\)-th stage, the spatial feature outputs \(\textbf{\textit{F}}^{\textit{l}}_i\) and \(\textbf{\textit{F}}^{\textit{r}}_i \in \mathbb{R}^{N_t \times c_i \times H_i W_i}\) are passed through a downsampling layer and fed into the next VSSD stage, generating refined feature maps. This iterative multi-stage process is defined as:  
\begin{equation} 
\setlength{\abovedisplayskip}{2pt}
\setlength{\belowdisplayskip}{0pt}
\textbf{\textit{F}}_{i+1} = \operatorname{VSSD}(\mathcal{D}(\textbf{\textit{F}}_{i})), 
\end{equation}
where \(\mathcal{D}(\cdot)\) represents the downsampling operation.

The cross-attention module facilitates interactions between features from the left and right views, generating fused features \(\textbf{\textit{F}}_c\). For the feature maps \(\textbf{\textit{F}}^{l}_a\) and \(\textbf{\textit{F}}^{r}_a\) obtained from the MSA stage, the module computes \textit{query} (\(\textbf{\textit{Q}}\)), \textit{key} (\(\textbf{\textit{K}}\)), and \textit{value} (\(\textbf{\textit{V}}\)) projections for both views. Queries from one view interact with keys from the other via a dot product, producing a cross-attention map \(\textbf{A}_{\text{cross}}\), which can be formulated as:
\begin{equation}
\setlength{\abovedisplayskip}{2pt}
\setlength{\belowdisplayskip}{0pt}
\textbf{A}_{\text{cross}} = \operatorname{Softmax}\left(\frac{\textbf{\textit{Q}}^l \left(\textbf{\textit{K}}^r\right)^T}{\sqrt{d_k}}\right),
\end{equation}
where \(\textbf{\textit{Q}}^l\) and \(\textbf{\textit{K}}^r\) are the query and key projections of the left and right views, respectively, and $d_k$
is the scaling factor. The attention map weights the value projection \(\textbf{\textit{V}}^{r}\), integrating information from the right view into the left, resulting in the fused features \(\textbf{\textit{F}}_c\).

To further enhance spatial representations, a transposed attention module is applied to prioritize channels within the fused features. This module computes  \textbf{\textit{Q}}, \textbf{\textit{K}}, and \textbf{\textit{V}} projections for \(\textbf{\textit{F}}_c\), reshaping the \(\textbf{\textit{Q}}\) and \(\textbf{\textit{K}}\) for dot-product interaction to produce a transposed attention map \(\textbf{A}_{\text{transposed}}\). The final spatial feature \(\tilde{\textbf{\textit{F}}}_V\) can be computed with a residual connection as:
\begin{equation}
\setlength{\abovedisplayskip}{0pt}
\setlength{\belowdisplayskip}{0pt}
\textbf{A}_{\text{transposed}} = \operatorname{Softmax}\left(\frac{\textbf{\textit{Q}}\textbf{\textit{K}}}{\sqrt{d_k'}}\right),
\end{equation}
\begin{equation}
\setlength{\abovedisplayskip}{0pt}
\setlength{\belowdisplayskip}{0pt}
\tilde{\textbf{\textit{F}}}_V = \textbf{\textit{W}}_p \cdot (\textbf{\textit{V}} \cdot \textbf{A}_{\text{transposed}}) + \textbf{\textit{F}}_c,
\vspace{-6pt}
\end{equation}
where \(\textbf{\textit{W}}_p\) is a learnable projection matrix.

\vspace{-4pt}
\subsection{Motion Feature Extractor} \vspace{-4pt}
While the aforementioned spatial feature extractor captures the spatial visual features of individual video frames, it neglects temporal information and fails to effectively integrate the underlying temporal visual features. Since motion distortions occur between frames (interframes), spatial features, which are derived from intraframes, are insufficient for representing such distortions. Therefore, motion features are essential for a comprehensive quality assessment of egocentric spatial videos.
To capture these motion features, we utilize the pretrained action recognition model SlowFast R50 \cite{feichtenhofer2019slowfast} as the motion feature extractor to capture the motion dynamics within the videos.
Given video chunks $\textbf{\textit{x}}^l$, $\textbf{\textit{x}}^r$ and the action recognition network $\mathcal{M}$, the motion features $\tilde{\textbf{\textit{F}}}_M$ can be represented as:
\begin{equation}
\setlength{\abovedisplayskip}{0pt}
\setlength{\belowdisplayskip}{0pt}
\tilde{\textbf{\textit{F}}}_M = \mathcal{M}(\textbf{\textit{x}}^l) \oplus \mathcal{M}(\textbf{\textit{x}}^r),
\end{equation}
where \(\oplus\) is the concatenation operator. 

\subsection{Semantic Feature Extractor} \vspace{-3pt}

Capturing high-level semantic information is crucial for models to enhance video understanding. To extract video global semantics, we sample \(N_t\) key frames \(\boldsymbol{y}^l\) and \(\boldsymbol{y}^r\) from the video and feed them into the CLIP visual encoder.
The input frames are processed by a patch embedding module \(\mathcal{P}\), which divides each frame into smaller patches and maps them into embedded feature maps. 
The feature maps are then passed through an \(m\)-layer transformer module \(\mathcal{T}^{(m)}\), which iteratively refines the features to emphasize relevant semantic information. Finally, the left and right view semantic features are concatenated into a unified representation  \(\tilde{\textbf{\textit{F}}}_S\):
\begin{equation}
\setlength{\abovedisplayskip}{3pt}
\setlength{\belowdisplayskip}{0pt}
\tilde{\textbf{\textit{F}}}_S = \mathcal{T}^{(m)}[\mathcal{P}(\textbf{\textit{y}}^l)] \oplus \mathcal{T}^{(m)}[\mathcal{P}(\textbf{\textit{y}}^r)],
\end{equation}
where \(\mathcal{P}\) denotes the patch embedding operation.

\vspace{-6pt}
\subsection{Features Fusion} \vspace{-4pt}
Previous quality assessment studies have employed a late fusion strategy \cite{zhu2023perceptual, sun2022deep}, in which multi-modal features are independently extracted and subsequently combined in the final stage to generate a comprehensive quality score. Adopting this approach, we utilize a multi-layer perceptron (MLP) to integrate the spatial features \(\tilde{\textbf{\textit{F}}}_V\), motion features \(\tilde{\textbf{\textit{F}}}_M\), and semantic features \(\tilde{\textbf{\textit{F}}}_S\), expressed as:
\begin{equation}
\setlength{\abovedisplayskip}{3pt}
\setlength{\belowdisplayskip}{0pt}
\hat{Q} = \operatorname{MLP}(\tilde{\textbf{\textit{F}}}_V \oplus \tilde{\textbf{\textit{F}}}_M \oplus \tilde{\textbf{\textit{F}}}_S),
\end{equation}
where \(\hat{Q}\) denotes the predicted video quality score and \(\oplus\) represents the concatenation operator.

\begin{table*}[t]
\centering
\vspace{-20pt}
\caption{Performance comparison of the state-of-the-art NR VQA models and the proposed ESVQAnet on our ESVQAD under three evaluation strategies. In the top-5 results for each criteria, the best result is marked in {\color[HTML]{FF0000} {\textbf{RED}}}, the second-best result is marked in {\color[HTML]{0070C0} {\textbf{BLUE}}}, and the remaining three are marked in \underline{Underlined}. } \vspace{-8pt}
\vspace{-2pt}
\label{1}
\setlength{\tabcolsep}{0.58em}
\scalebox{0.88}{
\begin{tabular}{l|cccc|cccc|cccc}
\toprule                             & \multicolumn{4}{c|}{left-view}          & \multicolumn{4}{c|}{right-view}         & \multicolumn{4}{c}{fusion-view}    \\
\midrule  Method  & SRCC$\uparrow$   & KRCC$\uparrow$   & PLCC$\uparrow$   & RMSE$\downarrow$   & SRCC$\uparrow$   & KRCC$\uparrow$   & PLCC$\uparrow$   & RMSE$\downarrow$   & SRCC$\uparrow$   & KRCC$\uparrow$   & PLCC$\uparrow$   & RMSE$\downarrow$   \\ \midrule
BMPRI \cite{bmpri}           & 0.1723 & 0.1197 & 0.1716 & 0.1769 & 0.1945 & 0.1354 & 0.1817 & 0.1766 & 0.2124 & 0.1463 & 0.2086 & 0.1756\\
BPRI\_LSSn \cite{pri}        & 0.4587 & 0.3223 & 0.4870 & 0.1568 & 0.4785 & 0.3369 & 0.5091 & 0.1546 & 0.4951 & 0.3488 & 0.5274 & 0.1526 \\
BPRI\_PSS  \cite{pri}        & 0.2229 & 0.1530 & 0.3541 & 0.1679 & 0.2357 & 0.1608 & 0.3290 & 0.1696 & 0.2322 & 0.1590 & 0.3276 & 0.1697 \\
BRISQUE \cite{mittal2012no}  & 0.4405 & 0.3061 & 0.4498 & 0.1604 & 0.4435 & 0.3088 & 0.4651 & 0.1590 & 0.4456 & 0.3110 & 0.4520 & 0.1602 \\
HOSA  \cite{hosa}            & 0.4355 & 0.3032 & 0.4479 & 0.1606 & 0.4556 & 0.3199 & 0.4674 & 0.1587 & 0.4649 & 0.3265 & 0.4792 & 0.1576 \\
NIQE  \cite{niqe}            & 0.3793 & 0.2620 & 0.4969 & 0.1558 & 0.3985 & 0.2759 & 0.5109 & 0.1544 & 0.4129 & 0.2864 & 0.5173 & 0.1537 \\
QAC   \cite{QAC}             & 0.3571 & 0.2407 & 0.3571 & 0.1677 & 0.3842 & 0.2620 & 0.3913 & 0.1653 & 0.4076 & 0.2789 & 0.4188 & 0.1631 \\
TLVQM  \cite{8742797}        & 0.4602 & 0.3274 & 0.5416 & 0.1328 & 0.5049 & 0.3615 & 0.5679 & 0.1325 & 0.5216 & 0.3751 & 0.5769 & 0.1302 \\
VIDEVAL \cite{tu2021ugc}     & 0.7070 & 0.5247 & 0.7362 & 0.0876 & 0.7101 & 0.5313 & 0.7388 & 0.0859 & 0.7202 & 0.5355 & 0.7537 & 0.0845 \\ \midrule
VSFA  \cite{li2019quality}   & 0.7997 & 0.6042 & 0.7728 & 0.0754 & 0.7981 & 0.6050 & 0.8030 & 0.0778 & 0.8128 & 0.6249 & 0.8290 & 0.0684 \\
FAST-VQA \cite{wu2022fast}    & \underline{0.8075} & \underline{0.6086} & 0.7897 & \underline{0.0654} & \underline{0.8007} & \underline{0.6074} & 0.7841 & \underline{0.0664} & \underline{0.8211} & \underline{0.6330} & 0.8114 & \underline{0.0626} \\
FasterVQA \cite{wu2022fasterquality} & 0.7810 & 0.5689 & 0.7772 & 0.0683 & 0.7980 & 0.6012 & 0.7758 & 0.0665 & 0.8019 & 0.6111 & 0.7722 & 0.0640 \\
Li \textit{et al.}  \cite{li2022blindly} & \underline{0.8109} & \underline{0.6235} & \underline{0.8300} & \underline{0.0634} & \underline{0.8148} & \underline{0.6263} & \underline{0.8327} & \underline{0.0620} & \underline{0.8295} & \underline{0.6375} & \underline{0.8493} & \underline{0.0598} \\
simpleVQA \cite{sun2022deep} & \underline{0.8190} & \underline{0.6204} & \underline{0.8139} & \underline{0.0651} & \underline{0.8156} & \underline{0.6146} & 0.8126 & \underline{0.0642} & {\color[HTML]{0070C0} {\textbf{0.8596}}} & {\color[HTML]{0070C0} {\textbf{0.6711}}} & {\color[HTML]{0070C0} {\textbf{0.8509}}} & {\color[HTML]{0070C0} {\textbf{0.0589}}} \\
DOVER \cite{wu2023exploring} & {\color[HTML]{0070C0} {\textbf{0.8293}}} & {\color[HTML]{0070C0} {\textbf{0.6339}}} & {\color[HTML]{0070C0} {\textbf{0.8313}}} & {\color[HTML]{0070C0} {\textbf{0.0615}}} & {\color[HTML]{0070C0} {\textbf{0.8307}}} & {\color[HTML]{0070C0} {\textbf{0.6381}}} & {\color[HTML]{0070C0} {\textbf{0.8329}}} & {\color[HTML]{0070C0} {\textbf{0.0610}}} & \underline{0.8420} & \underline{0.6450} & 0.8360 & \underline{0.0621} \\
MBVQA\_b \cite{wen2024modular}  & 0.7697 & 0.5766 & 0.7967 & 0.0766 & 0.7740 & 0.5846 & 0.8012 & 0.0767 & 0.7957 & 0.6116 & 0.8235 & 0.0705 \\
MBVQA\_s \cite{wen2024modular}  & 0.7655 & 0.5761 & 0.8000 & 0.0760 & 0.7758 & 0.5870 & 0.8080 & 0.0764 & 0.7956 & 0.6137 & 0.8288 & 0.0741 \\
MBVQA\_t \cite{wen2024modular}  & 0.7814 & 0.5906 & 0.8095 & 0.0751 & 0.7836 & 0.5971 & \underline{0.8159} & 0.0755 & 0.8126 & 0.6258 & \underline{0.8361} & 0.0656 \\
MBVQA\_st \cite{wen2024modular} & 0.7787 & 0.5892 & \underline{0.8105} & 0.0740 & 0.7873 & 0.6038 & \underline{0.8170} & 0.0740 & 0.8172 & 0.6329 & \underline{0.8428} & 0.0652 \\
\rowcolor{mygray} ESVQAnet         & {\color[HTML]{FF0000}{\textbf{0.8447}}} & {\color[HTML]{FF0000}{\textbf{0.6555}}} & {\color[HTML]{FF0000}{\textbf{0.8552}}} & {\color[HTML]{FF0000}{\textbf{0.0571}}} & {\color[HTML]{FF0000}{\textbf{0.8422}}} & {\color[HTML]{FF0000}{\textbf{0.6545}}} & {\color[HTML]{FF0000}{\textbf{0.8491}}} & {\color[HTML]{FF0000}{\textbf{0.0581}}} & {\color[HTML]{FF0000}{\textbf{0.8867}}} & {\color[HTML]{FF0000}{\textbf{0.7070}}} & {\color[HTML]{FF0000}{\textbf{0.8733}}} & {\color[HTML]{FF0000}{\textbf{0.0546}}} \\
\bottomrule
\end{tabular} 
}
\vspace{-20pt}
\end{table*}

\vspace{-4pt}
\section{Experiment}
\vspace{-4pt}
\subsection{Experimental Setup}
\vspace{-2pt}
\subsubsection{Compared Methods}
To evaluate the performance of ESVQAnet, we compare it with 16 state-of-the-art no-reference (NR) VQA models designed for traditional 2D videos, which can be categorized into two groups:
\begin{itemize}
\item \textbf{Hand-crafted models:} BRISQUE \cite{mittal2012no}, BMPRI \cite{bmpri}, NIQE \cite{niqe}, HOSA \cite{hosa}, BPRI-PSS \cite{pri}, BPRI-LSSn \cite{pri}, QAC \cite{QAC}, VIDEVAL \cite{tu2021ugc}, and TLVQM \cite{8742797}.
\item \textbf{Deep learning-based models:} VSFA \cite{li2019quality}, simpleVQA \cite{sun2022deep}, FAST-VQA \cite{wu2022fast}, FasterVQA \cite{wu2022fasterquality}, Li \textit{et al.} \cite{li2022blindly}, DOVER \cite{wu2023exploring}, and ModularBVQA (MBVQA) \cite{wen2024modular}.
\end{itemize}

\subsubsection{Evaluation Metrics}

We assess the correlation between predicted scores and MOSs using four evaluation criteria, \textit{i.e.,} Spearman rank correlation coefficient (SRCC), Pearson linear correlation coefficient (PLCC), Kendall’s rank correlation coefficient (KRCC), and root mean squared error (RMSE). SRCC, PLCC, and KRCC measure the prediction monotonicity, while RMSE measures the prediction accuracy. 
Before computing these criteria, predicted scores are normalized using a five-parameter logistic function:
\begin{equation}
\setlength{\abovedisplayskip}{3pt}
\setlength{\belowdisplayskip}{2pt}
   \hat{y}=\beta_1\left(0.5-\frac{1}{1+e^{\beta_2\left(y-\beta_3\right)}}\right)+\beta_4 y+\beta_5, 
\end{equation}
where $\left\{\beta_i \mid i=1,2, \ldots, 5\right\}$ are parameters to be fitted, $y$ and $\hat{y}$ indicate the predicted scores and the mapped scores. 

\subsubsection{Implementation Details}

For each VQA model, we evaluate the quality of egocentric spatial video using three strategies, \textit{i.e.,} single-view predictions by the left-view and right-view video sequences separately, and a fusion-view prediction combining both views.
For traditional hand-crafted benchmark VQA models, the quality score of fusion-view prediction is simply the average of left and right view predictions. For deep learning-based benchmark models, we train and test them on the ESVQAD database, which is split into training and testing sets at a 4:1 ratio. The training parameters are consistent with those in the officially released versions. 
Since spatial videos cannot be directly input into these models, we fine-tune their architectures to support dual-video inputs (6 channels).
Our proposed ESVQAnet extracts multi-scale features through three stages with \(N_1=3\), \(N_2=4\), \(N_3=21\) VSSD blocks and a final stage with \(N_4=5\) MSA blocks. In single-view evaluation, we omit the branch for the other view, along with the feature concatenation and cross-attention modules, while keeping the rest of the model unchanged. Additionally, we pretrain the spatial extractor on the ESIQAD database (500 real-scene egocentric spatial images) \cite{zhu2024esiqa}, using transfer learning to improve the model's sensitivity to quality variations. 
To ensure comparability, we use the same train/test splits as the benchmark models. Furthermore, we employ PLCC as the optimization goal, which combines the benefits of both regression and learning-to-rank formulations  \cite{wen2024modular}.

\vspace{-6pt}
\subsection{Performance Analysis}
\vspace{-3pt}
We evaluate the aforementioned 16 state-of-the-art models and our proposed model on the ESVQAD. The experimental results are demonstrated in Table \ref{1}, showing that ESVQAnet outperforms all benchmark models across the three evaluation strategies, demonstrating its ability to effectively capture and integrate spatial and temporal features of egocentric spatial videos. Specifically, ESVQAnet achieves a 1.54\% improvement in SRCC and a 2.39\% improvement in PLCC over the best benchmark model DOVER \cite{wu2023exploring} for left-view evaluation, a 1.15\% improvement in SRCC and a 1.62\% improvement in PLCC over the best benchmark model DOVER \cite{wu2023exploring} for right-view evaluation, and a 2.71\% improvement in SRCC and a 2.24\% improvement in PLCC  over the best benchmark model simpleVQA \cite{sun2022deep} for fusion-view evaluation.
All models show improved performance in fusion-view evaluation compared to single-view evaluation, emphasizing that disparity information from the left and right views enhances the model’s perception of the 3D stereoscopic effect in immersive environments. Additionally, most models perform slightly better with right-view evaluation than left-view evaluation, suggesting that the right view may offer richer spatial and dynamic information.

\begin{table}[t]
\vspace{-20pt}
\centering
\caption{Ablation study on the three feature extractors in our ESVQAnet. \(\tilde{\textbf{\textit{F}}}_V\), \(\tilde{\textbf{\textit{F}}}_M\), and \(\tilde{\textbf{\textit{F}}}_S\) refers to the spatial features, motion features, and semantic features, respectively. All numbers are presented in the SRCC / PLCC format. Best performances are indicated with \textbf{BOLD}.}
\vspace{-8pt}
\label{ablation}
\setlength{\tabcolsep}{0.6em}
\scalebox{0.90}{
\begin{tabular}{c c c|c c c}
\toprule
 \(\tilde{\textbf{\textit{F}}}_V\) & \(\tilde{\textbf{\textit{F}}}_M\) & \(\tilde{\textbf{\textit{F}}}_S\) & left-view & right-view & fusion-view \\
\midrule
\midrule
$\checkmark$ &  & & 0.8010 / 0.7964 & 0.8091 / 0.8055 & 0.8527 / 0.8401 \\ 
 &  $\checkmark$ & & 0.6553 / 0.6704 & 0.6425 / 0.6676 & 0.7466 / 0.7629 \\
 & &  $\checkmark$ & 0.4742 / 0.5614 & 0.5010 / 0.6281 & 0.5597 / 0.6578 \\ 
$\checkmark$ &  & $\checkmark$ & 0.8120 / 0.8042 & 0.8167 / 0.8029 & 0.8668 / 0.8584 \\
$\checkmark$ & $\checkmark$  & &  0.8228 / 0.8175 & 0.8261 / 0.8199 & 0.8713 / 0.8688\\
& $\checkmark$ & $\checkmark$ & 0.6764 / 0.6950 & 0.7134 / 0.7204 & 0.7675 / 0.7797 \\
 $\checkmark$ & $\checkmark$ & $\checkmark$ & \baseline{\textbf{0.8447} / \textbf{0.8552}} & \baseline{\textbf{0.8422} / \textbf{0.8491}} & \baseline{\textbf{0.8867} / \textbf{0.8733}} \\
\bottomrule
\end{tabular}
}
\vspace{-12pt}
\end{table}

\begin{table}[!t]
\centering
\caption{Ablation study on \textbf{backbones for the spatial feature extractor}. Results are shown as SRCC / PLCC, with the best performances in \textbf{BOLD}.}
\vspace{-8pt}
\label{tab:backbone}
\setlength{\tabcolsep}{0.55em}
\scalebox{0.89}{
\vspace{-5pt}
\begin{tabular}{lccc}
    \toprule
  \rule{-2pt}{2pt}  Backbone & left-view  & right-view & fusion-view  \\
    \midrule
    RN50-ImageNet & 0.8190 / 0.8260 & 0.8202 / 0.8307 & 0.8602 / 0.8610 \\
    ViT-ImageNet & 0.8211 / 0.8207 & 0.8288 / 0.8337 & 0.8691 / 0.8707 \\
    VSSD-ImageNet & 0.8329 / 0.8475 & 0.8409 / 0.8433 &  0.8757 / \textbf{0.8737}\\
    VSSD-ESIQAD & \baseline{\textbf{0.8447} / \textbf{0.8552}} & \baseline{\textbf{0.8422} / \textbf{0.8491}} & \baseline{\textbf{0.8867} / 0.8733} \\
    \bottomrule
\end{tabular}
}
\vspace{-12pt}
\end{table}

\begin{table}[!t]
\centering
\caption{Ablation study on \textbf{backbones for the motion feature extractor}. Results are shown as SRCC / PLCC, with the best performances in \textbf{BOLD}.} \vspace{-8pt}
\label{tab:motion}
\setlength{\tabcolsep}{0.55em}
\scalebox{0.89}{
\vspace{-5pt}
\begin{tabular}{lccc}
    \toprule
  \rule{-2pt}{2pt} Backbone  & left-view  & right-view & fusion-view  \\
\midrule
R(2+1)D  &   0.8096 / 0.8152 & 0.8108 / 0.8155 & 0.8406 / 0.8441  \\
MViT &   0.8181 / 0.8405 & 0.8144 / 0.8322 & 0.8581 / 0.8593 \\
Slow  &   0.8348 / 0.8426 & 0.8420 / 0.8419 & 0.8653 / 0.8362  \\
SlowFast & \baseline{\textbf{0.8447} / \textbf{0.8552}} & \baseline{\textbf{0.8422} / \textbf{0.8491}} & \baseline{\textbf{0.8867} / \textbf{0.8733}} \\
\bottomrule
\end{tabular}
}
\vspace{-15pt}
\end{table}

\begin{table}[t]
\vspace{-20pt}
\centering
\caption{Ablation study on \textbf{quality regressors}. Results are shown as SRCC / PLCC, with the best performances in \textbf{BOLD}.} \vspace{-8pt}\label{tab:regressor}
\setlength{\tabcolsep}{0.55em}
\scalebox{0.88}{
\vspace{-5pt}
\begin{tabular}{lccc}
    \toprule
  \rule{-2pt}{2pt}   & left-view  & right-view & fusion-view  \\
    \midrule
    GRU & 0.8405  / 0.8512  & 0.8389 / 0.8417  & 0.8820  / 0.8714  \\
    Transformer & 0.8408 / 0.8526  & 0.8352  / 0.8443  & 0.8815 / 0.8706 \\
    MLP & \baseline{\textbf{0.8447} / \textbf{0.8552}} & \baseline{\textbf{0.8422} / \textbf{0.8491}} & \baseline{\textbf{0.8867} / \textbf{0.8733}} \\
    \bottomrule
\end{tabular}}
\vspace{-14pt}
\end{table}

\begin{table}[ht]
\centering
\caption{Ablation study on \textbf{loss functions}. Results are shown as SRCC / PLCC, with the best performances in \textbf{BOLD}.} \vspace{-8pt}\label{tab:loss}
\setlength{\tabcolsep}{0.55em}
\scalebox{0.89}{
\begin{tabular}{lccc}
    \toprule
  \rule{-2pt}{2pt}   & left-view  & right-view & fusion-view  \\
    \midrule
    $\ell_1$ & 0.8223  / 0.8312  & 0.8235  / 0.8348  & 0.8648 / 0.8808  \\
    $\ell_2$ & 0.8215  / 0.8275  & 0.8199 / 0.8361  & 0.8556  / \textbf{0.8756} \\
    PLCC & \baseline{\textbf{0.8447} / \textbf{0.8552}} & \baseline{\textbf{0.8422} / \textbf{0.8491}} & \baseline{\textbf{0.8867} / 0.8733} \\
    \bottomrule
\end{tabular}}
\vspace{-14pt}
\end{table}


\vspace{-6pt}
\subsection{Ablation Study}
\vspace{-2pt}

\begin{table}[h!]
\centering
\caption{Performance of the state-of-the-art models and the proposed ESVQAnet on the KoNViD-1k, YouTube-UGC, LBVD, and LIVE-YT-Gaming databases. All performance is presented in the SRCC / PLCC format. The best-performing model is highlighted in \textbf{BOLD} in each column. } \vspace{-5pt}
\vspace{-2pt}
\label{traditionalVQA}
\setlength{\tabcolsep}{0.45em}
\scalebox{0.77}{
\begin{tabular}{c|l|cccc}
\toprule  & Database &  KoNViD-1k\cite{hosu2017konstanz}  & YouTube-UGC\cite{wang2019youtube}  &  LBVD\cite{chen2019qoe}   & LIVE-YT-G\cite{yu2022subjective}  \\
\midrule \multirow{4}{*}{\rotatebox{90}{IQA}}&   NIQE \cite{niqe}& 0.542 / 0.553 & 0.238 / 0.278 & 0.327 / 0.387 & 0.280 / 0.304\\
 &  BRISQUE \cite{mittal2012no} & 0.657 / 0.658 & 0.382 / 0.395 & 0.435 / 0.446 & 0.604 / 0.638 \\
 &  VGG19 \cite{vgg} & 0.774 / 0.785 & 0.703 / 0.700 & 0.676 / 0.673 & 0.678 / 0.658  \\
 &  ResNet50 \cite{resnet} & 0.802 / 0.810 & 0.718 / 0.710 & 0.715 / 0.717 & 0.729 / 0.768  \\
\midrule  
\multirow{10}{*}{\rotatebox{90}{VQA}}& V-BLIINDS \cite{saad2014blind} & 0.710 / 0.704 & 0.559 / 0.555 & 0.527 / 0.558 & 0.357 / 0.403  \\
 &  TLVQM \cite{8742797} & 0.773 / 0.769 & 0.669 / 0.659 & 0.614 / 0.590 & 0.748 / 0.756 \\
 &  VIDEVAL \cite{tu2021ugc} & 0.783 / 0.780 & 0.779 / 0.773 & 0.707 / 0.697 & 0.807 / 0.812  \\
 &  VSFA \cite{li2019quality} & 0.773 / 0.775 & 0.724 / 0.743 & 0.622 / 0.642 & 0.776 / 0.801 \\
 &  Li \textit{el al.} \cite{li2022blindly} & 0.836 / 0.834 & 0.831 / 0.819 & - / - & - / - \\
 &  simpleVQA \cite{sun2022deep} & 0.856 / 0.860 & 0.847 / 0.856 & 0.844 / 0.846 & 0.861 / 0.866 \\
 &  FAST-VQA \cite{wu2022fast} & 0.891 / 0.892 & 0.855 / 0.852 & 0.804 / 0.809 & 0.869 / 0.880 \\
  &  FasterVQA \cite{wu2022fast} & 0.895 / 0.898 & 0.863 / 0.859 & 0.813 / 0.837 & 0.867 / 0.872 \\
 &  DOVER \cite{wu2023exploring} & 0.909 / 0.906 & 0.890 / 0.891 & - / - & - / - \\
 \rowcolor{mygray} &  ESVQAnet &  \textbf{0.913} / \textbf{0.908} & \textbf{0.893} /  \textbf{0.899} & \textbf{0.869} / \textbf{0.873} &  \textbf{0.888} / \textbf{0.905} \\
\bottomrule
\end{tabular}
}    
\vspace{-18pt}
\end{table}

To validate the impact of each component in ESVQAnet, we conduct ablation studies on the three feature extractors in our ESVQAnet in Table \ref{ablation}, backbones for spatial feature extractor in Table \ref{tab:backbone}, backbones for motion feature extractor in Table \ref{tab:motion}, quality regressors in Table \ref{tab:regressor}, and loss functions in Table \ref{tab:loss}.
The results in Table \ref{ablation} show that combining all three feature extractors achieves the best performance across left, right, and fusion-view evaluations, highlighting the complementary role of semantic and motion features in enhancing spatial feature-based quality assessment for egocentric spatial videos.
The ablation studies in Table \ref{tab:backbone} show that VSSD outperforms both ResNet-50 \cite{resnet} and ViT \cite{dosovitskiy2020image} pretrained on ImageNet-1K, demonstrating that our spatial feature extractor benefits from a more robust VSSD backbone, particularly the pretrained version on ESIQAD \cite{zhu2024esiqa}. 
Additionally, the results in Table \ref{tab:motion} demonstrate that SlowFast effectively captures time-domain information, outperforming MViT \cite{li2022mvitv2} and R(2+1)D \cite{manasa2016optical} on our ESVQAD. Even when only Slow features are extracted, the model performs satisfactorily. 
The results in Table \ref{tab:regressor} indicate that the choice of quality regressors for feature fusion has minimal impact on quality prediction. More complex models like GRU and transformer offer no improvement over MLP.
The results in Table \ref{tab:loss} show that using PLCC as the loss function significantly improves performance compared to \(\ell_1\)-norm and \(\ell_2\)-norm induced metrics.

\vspace{-3pt}
\subsection{Generalization Ability Validation}
\vspace{-3pt}

Since the generalization ability in the traditional VQA task is crucial for egocentric spatial VQA models, we use LSVQ \cite{ying2021patch} as the large database and choose four small databases representing diverse scenarios, \textit{i.e.,} LBVD \cite{chen2019qoe}, KoNViD-1k ~\cite{hosu2017konstanz}, LIVE-YT-Gaming~\cite{yu2022subjective} and YouTube-UGC \cite{wang2019youtube} to evaluate the performance of ESVQAnet and the existing state-of-the-art VQA models. 
As Table \ref{traditionalVQA} shows, the ESVQAnet achieves competitive performance among the state-of-the-art VQA models on all traditional VQA scenarios, demonstrating strong robustness and generalization ability.

\section{Conclusion}
In this work, we explore the embodied perceptual quality assessment problem of egocentric spatial videos. The first large-scale egocentric video quality assessment database, ESVQAD, is established, which consists of 600 egocentric spatial videos with corresponding subjective quality ratings. To advance this field, we benchmark the performance of 16 state-of-the-art VQA models and highlight their limitations for egocentric spatial videos. 
We also propose ESVQAnet, a novel multi-dimensional binocular feature fusion model that incorporates spatial, motion, and semantic features binocularly. Extensive experimental results demonstrate that ESVQAnet achieves state-of-the-art performance on both ESVQAD and other VQA databases.
Our work establishes a foundation for future research in egocentric spatial video quality assessment, and we hope the research can extend its impact to areas of embodied intelligence, such as telerobotics, telemedicine, \textit{etc.}, and facilitate their development.

\vspace{-2pt}
\bibliographystyle{IEEEbib}
\bibliography{main}

\end{document}